# Model-based gait recognition using graph network on very large population database

Zhihao Wang[1], Chaoying Tang [1] (✉),

[1] College of Automation Engineering, Nanjing University of Aeronautics and Astronautics, Nanjing 211106, China
cytang@nuaa.edu.cn

**Abstract.** At present, the existing gait recognition systems are focusing on developing methods to extract robust gait feature from silhouette images and they indeed achieved great success. However, gait can be sensitive to appearance features such as clothing and carried items. Compared with appearance-based method, model-based gait recognition is promising due to the robustness against these variations. In recent years, with the development of human pose estimation, the difficulty of model-based gait recognition methods has been mitigated. In this paper, to resist the increase of subjects and views variation, local features are built and a siamese network is proposed to maximize the distance of samples from the same subject. We leverage recent advances in action recognition to embed human pose sequence to a vector and introduce Spatial-Temporal Graph Convolution Blocks (STGCB) which has been commonly used in action recognition for gait recognition. Experiments on the very large population dataset named OUMVLP-Pose and the popular dataset, CASIA-B, show that our method archives some state-of-the-art (SOTA) performances in model-based gait recognition. The code and models of our method are available at https://github.com/timelessnaive/Gait-for-Large-Dataset after being accepted.

## 1 Introduction

Gait is one of the most popular behavioral biometrics in the world because it has unique advantages compared with face, iris, palm print, etc. Gait features can be captured at a long distance and are hard to disguise, and consequently, gait recognition technology has been added to the repertoire of tools available for crime prevention and forensic identification. Most of the current methods [1-3] usually take the silhouette sequence extracted from the video as the input. In general, there are three steps in appearance-based methods: silhouette extraction, feature learning, and distance measurement. However, most of the studies do not include a very specific approach to silhouette extraction. The existing silhouette extraction approaches such as background subtraction [4] fail to perform well in cluttered and rapidly-changing scenarios. In addition, the silhouette contains not only gait information but also other appearance cues such as body shape and hairstyle. The complexity of information weakens its robustness especially when appearance changed. On the contrary, the model-based approach has achieved more attention since current pose estimation algorithms are robust against changing carried items and clothing. Most model-based methods focus on extracting static or dynamic features from skeletons such as the distance between joints, the angle of limbs [5-6]. While these features describe global information of different dimensions, the local features of joints and bones are ignored. Moreover, due to the success of graph networks in action recognition, some research [7] introduced the graph network into gait recognition, while the gait sequences are more similar than actions, which means that it is more difficult to classify or match. This problem is more prominent for the large dataset. Nowadays, Gait recognition based on skeleton achieved high recognition accuracy (more than 90% [2]) on a smaller dataset (such as CASIA-B dataset with 128 subjects), while the accuracy decreases heavily (lower than 40%) when the number of subjects increases (OUMVLP-Pose dataset with 10307 subjects). The comparison of such two datasets is shown in Table 1. For such a large dataset, we intend to build the discerning features to describe the local gesture and siamese structure to maximum the distance between same subject.

In this paper, we take the pose sequences from pose estimation and take the joint, acceleration, and bone features as the input. Furthermore, we adopt the siamese structure and take the AGCN network which is commonly used in skeleton-based action recognition as the embedding network. Concretely, the AGCN network is the stack of Spatial-Temporal Graph Convolution Blocks (STGCB). Then, we use the Euclidean distance of the corresponding feature vectors to measure the similarity between gallery and probe. Our experiments on the OUMVLP-Pose gait dataset have proved the effectiveness of the proposed method. More specifically, we achieved an average rank-1 recognition rate of 38.07% on the OpenPose part and 58.02% on the AlphaPose part which achieves the state-of-the-art performance. We also do experiments on the CASIA-B dataset, a widely used gait dataset, and still achieve competitive results against the appearance-based method.

Our contributions can be summarized as follows:

(1) We introduced graph neural network into gait recognition, and we build the joint, acceleration and bone's angle features to enrich the input of the network.

(2) To enhance the robustness of gait recognition models to skeletons, we adopt the siamese structure and use a supervised contrastive learning loss function to increase the distance between different classes.

(3) Our empirical experiments show state-of-the-art results compared to the current model-based approaches on the very large population dataset named OUMVLP-Pose and the popular dataset, CASIA-B.

**Table 1.** Existing major gait databases with multi-view images

| Name | #subjects | #views | View range | #seqs | Year |
| --- | --- | --- | --- | --- | --- |
| CASIA-B [8] | 124 | 11 | 0-180 | 13640 | 2006 |
| OUMVLP-Pose [9] | 10307 | 14 | 0-90, 90-270 | 20614 | 2020 |

## 2   Related work

*A.   Features representation in gait*

The current work of gait recognition can basically be categorized into two types. One type is the appearance-based method. Early approaches of this part proposed to encode a gait sequence into a single image, i.e., Gait Energy Image (GEI) [10]. A sequence of silhouettes can represent useful gait features such as speed, cadence, leg angles, gait cycle time, step length, stride length, and the ratio between swing and stance phases [11-12]. It can also be processed to extract motion data, like using optical flow calculation [13-14]. Nonetheless, to the limitation of gait silhouettes, these features are more sensitive to changes in the appearance of the individuals, such as different clothing and carrying conditions.

The other type is the model-based method. Current work tends to use skeleton to represent the body. Skeleton can be captured using depth-sensing cameras [15] or alternatively estimated using pose estimation methods [16]. Static and dynamic features such as stride length, speed, distances, and angles between joints can be obtained from connected body joints in the form of skeletons [17]. Wang et al. [18] divided the human body into 14 parts and used joint-angle trajectories in each part for identifying individuals. They also combined Procrustes shape analysis for improving recognition rate.

Alireza et al. [19] shows that over 87% of the methods available are based on global feature representations, where the deep features are learned by considering the gait information as a whole. In this paper, we intend to build dynamic local features to describe walking pattern more clearly. In our local feature, inter-frame and intra-frame features are extracted from original pose sequences and we use siamese structure to decrease the possibility of false probe-gallery matching.

*B.   Networks on pose sequences*

Convolutional neural networks (CNNs) have been used mostly for gait recognition. Liao et al. [20] proposed a pose-based temporal-spatial network (PTSN) to extract the temporal-spatial features, which effectively improves the performance of gait recognition. In addition, they use the Long Short-Term Memory (LSTM) and CNN to extract the temporal and the spatial features respectively from static gait pose frames. With the success in the similar field like action recognition, graph convolutional networks (GCNs) have been recently developed to extend CNNs. GCNs can jointly model both the structural information and temporal relationships available in a gait sequence. For example, the gait features of [21] were extracted from video sequences by a spatio-temporal graph network. Then they using a joint relationship learning scheme to obtain the features. As the result, the gait features were embedded into a more discriminative subspace to describe walking pattern information. However, even the GCN is introduced into gait recognition, its power in gait recognition had been underestimated at present.

Using deep learning to encode pose sequences from special hardware like deep sensors or Kinect is widely adopted by researchers. For example, Kastaniotis et al. [22] used skeleton data from a single Kinect sensor instead of multiple synchronous cameras in [23]. Some researches show the advancement of encoding skeleton extracted by pose estimation based on deep learning. Recently Liao et al. [24] used pose estimation by deep learning to recover human skeleton models. They also converted 2D pose data to 3D for view-invariant feature extraction. We believe that model-based methods will be promoted greatly with the development of pose estimation and deep learning.

## 3  Skeleton-based gait recognition

In this section, we describe our method for learning discriminative information from sequences of human poses.

### 3.1  Pipeline overview

The overview of our model is presented in Fig 1, and the main components of our framework are:

**Feature Extraction module,** $Extra(\cdot)$. We extract the joint, acceleration, and bone features to utilize the second-order information. Before the extraction, we apply augmentation on every input skeleton sequence $x$ twice to generate two copies, $\tilde{x}_1, \tilde{x}_2$. For each copy, $\tilde{x}$, we extract features like $\tilde{X} = [f_{joint}(\tilde{x}), f_{acceleration}(\tilde{x}), f_{bone}(\tilde{x})]$ and concatenated two extractions, $X = [\tilde{X}_1 | \tilde{X}_2]$, each of which represents a different view of the data and contains some subset of the information in the original sample. Details of the features will be given in Section 3.2.

**Embedding Network,** $Embed(\cdot)$, which maps $x$ to a representation vector, $r = Ebd(x) \in R^{D_E}$. Both augmented samples are separately input to the same embedding network, resulting in a pair of representation vectors. $r$ is normalized to the unit hypersphere in $R^{D_E}$ ($D_E = 256$ in all our experiments in the paper). Here, we use siamese structure and take AGCN as the embedding network.

**Match module,** $Match(\cdot)$, which adopts the Euclidean distance of the corresponding feature vectors $r$ to measure the similarity between gallery and probe.

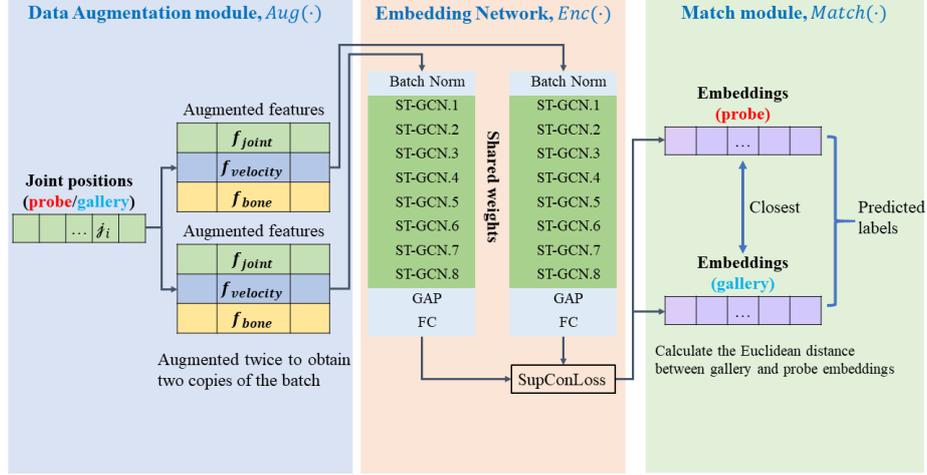

**Fig. 1.** Overview of the pipeline.

### 3.2 Feature Extraction module

To enrich the features of data, we constructed two types of features: intra-frame features and inter-frame features. Intra-frame features such as the relative position of joints reflect static information while inter-frame features like the acceleration of joint, etc., describe dynamic information of coherent actions. Since different walking speeds will affect the inter-frame features, we use different sampling periods to simulate fast and slow walking. The overview of our three features is shown in Fig.2.

A human skeleton graph is denoted as $\mathcal{G} = (\mathcal{J}, \mathcal{B})$, where $\mathcal{J} = \{j_1, ..., j_N\}$ is the set of $N$ nodes representing joints and $\mathcal{B}$ is the set of edges representing bones captured by an adjacency matrix $A \in \mathbb{R}^{N \times N}$. And $A_{i,j} = 1$ represents that an edge connects joint $v_i$ and joint $v_j$. The $i_{th}$ joint in frame $t$ is denoted as $j_i^t$ and the coordinates are $(x_i^t, y_i^t)$.

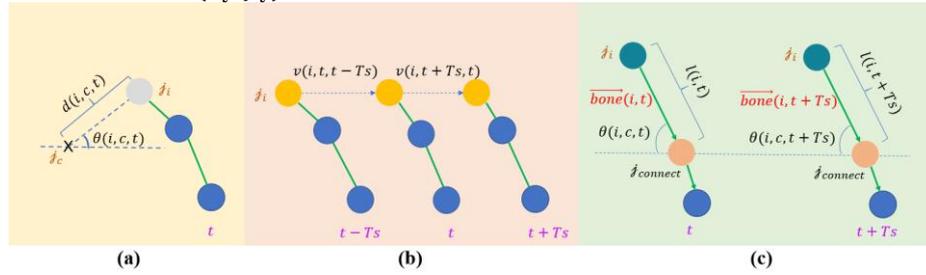

**Fig. 2.** The three local features.
(a) Joint features. (b) Acceleration features. (c) Bone features.

**Intra-frame features: Joint features.**

The joint features $f_{joint}$ computes relative positions of a joint to the center joint. The average coordinate $(x_c^t, y_c^t)$ of all joints in the skeleton at a frame is treated as its center. For convenience, we take $c$ as the index in the following paragraph.
By computing the positions of a joint to its neighbor joints. The $f_{joint}$ describes the biometric features of the body such as scale and shape. The joint feature between joint $i$ and $j$ in frame $t$ is a vector that contains Euclidean distance $d(i,j,t)$ and the angle $\theta(i,j,t)$. The $f_{joint}$ of a gait sequence calculates the joint features among all frames.

$$x_c^t = \frac{\sum_{k=0}^{N} x_k^t}{N}$$

$$y_c^t = \frac{\sum_{k=0}^{N} y_k^t}{N}$$

$$d(i,j,t) = \sqrt{\left(x_i^t - x_j^t\right)^2 + \left(y_i^t - y_j^t\right)^2}$$

$$\theta(i,j,t) = arctan\left(\frac{y_i^t - y_j^t}{x_i^t - x_j^t}\right)$$

$$f_{joint}(i,t) = [d(i,c,t), \theta(i,c,t)]$$

$$f_{joint}(t) = f(i,t)|_{i=0}^{N}$$

$$f_{joint} = f_{joint}(t)|_{t=0}^{T}$$

**Inter-frame features: Acceleration features.**
Compared to velocity, acceleration is a better feature which describes a person's muscle strength and walking habits more vividly. We use the position variation of the same joint between two adjacent frames to express its velocity, and acceleration is calculated from the velocity vector size and angle variation in three consecutive frames. Different sampling periods are considered in $f_{acceleration}$ to enhance the robustness. For the $t_{th}$ joint $i$, the velocity feature $v(i,t,t-Ts)$ is defined as the difference between the coordinates of the joint point numbered $i$ between $t$ and $t-Ts$ frames, where Ts is the sampling period. The acceleration feature $a(i,t,Ts)$ is the difference between the velocity features in adjacent frames. Furthermore, $f(i,t,Ts)$ include three frames of $t-Ts$, $t$ and $t+Ts$, and $f_{acceleration}$ includes two acceleration features with sampling periods of 1 and 2. For joints without adjacent frames, the missing features are filled with zeros.

$$v(i,t_1,t_0) = (x_i^{t_1} - x_i^{t_0}, y_i^{t_1} - y_i^{t_0})$$

$$a(i,t,Ts) = v(i,t+Ts,t) - v(i,t,t-Ts)$$

$$f(i,t,Ts) = [a(i,t-Ts,Ts), a(i,t,Ts), a(i,t+Ts,Ts)]$$

$$f_{acceleration}(i,t) = [f(i,t,1), f(i,t,2)]$$

$$f_{acceleration} = (f_{acceleration}(i,t)|_{t=1}^{T})|_{i=0}^{N}$$

**Inter-frame features: Bone features.**
Bone features: For bone features, we treat each bone as a vector. By calculating the variation in the angle of bones and length between adjacent frames, the movement of the bones is described. Different sampling periods are also used to simulate different walking speeds. For the $t^{th}$ frame, $f_{bone}$ is defined as the sequence of distances and angles as below. The joint which connects with joint $i$ is denoted as $(x_{connect}^t, y_{connect}^t)$. For joints that are connected with more than one joint, we take the closer one. The bone vector is defined as $bone(i,t) = (x_i^t - x_{connect}^t, y_i^t - y_{connect}^t)$ and the length of bone is $l(i,t)$, while $\theta(i,t)$ described the angle of the bone. For the joints that lack adjacent frames, the missing features are filled with zero.

$$bone(i,t) = (x_i^t - x_{connect}^t, y_i^t - y_{connect}^t)$$
$$l(i,t) = \|bone(i,t)\|$$
$$\theta(i,t) = \arctan\left(\frac{y_i^t - y_{connect}^t}{x_i^t - x_{connect}^t}\right)$$
$$\Delta l(i,t,Ts) = l(i,t) - l(i,t-Ts)$$
$$\Delta \theta(i,t,Ts) = \theta(i,t) - \theta(i,t-Ts)$$
$$g_1(i,t) = [\Delta l(i,t,1), \Delta \theta(i,t,1)]$$
$$g_2(i,t) = [\Delta l(i,t,2), \Delta \theta(i,t,2)]$$
$$G(i,t,Ts) = [l(i,t), \theta(i,t), g_1(i,t), g_2(i,t)]$$
$$f_{bone}(i,t) = [G(i,t,1), G(i,t,2)]$$
$$f_{bone} = (f_{bone}(i,t)|_{t=1}^T)|_{i=0}^N$$

Finally, our features consist of three local features, which is defined as

$$\tilde{X}_k = [f_{joint}(\tilde{x}_k), f_{acceleration}(\tilde{x}_k), f_{bone}(\tilde{x}_k)], \tag{4}$$

where k is the index of augment copies. The final output tensor of feature extraction module is $X = [\tilde{X}_1 | \tilde{X}_2]$.

### 3.3 Embedding Network

#### 3.3.1 Spatial-Temporal Graph Convolutional Networks (ST-GCN) block

**Graph conventional networks.** Gait can be seen as a sequence of graphs. In our program, the node feature set is represented by a feature tensor $X \in \mathbb{R}^{B \times T \times N \times F \times C}$, where $B$ is the index of batch views and $x_{t,n,f} \in X$ is the $C$ dimensional feature vector for node $v_{n,f}$ in $f_{th}$ feature at time $t$ over total $T$ frames. In our paper, every node $v_{n,f}$ in feature tensor $X$ consists of 6 scales.

The inputs to the network are feature $X$ and graph structure $A$. The graph structure $A$ is an adjacency matrix which represents the information of connections.

The layer-wise update rule of Graph Convolutional Networks (GCNs) can be applied to features at time $t$ as:

$$X_t^{l+1} = \sigma(\tilde{D}^{-\frac{1}{2}} \tilde{A} \tilde{D}^{-\frac{1}{2}} X_t^l W^l) \tag{5}$$

where $\tilde{A} = A + I$ is identity features, $I$ is an identity matrix, $\tilde{A}$ is the skeleton graph to keep identity features, $W^l \in \mathbb{R}^{C_l \times C_{l+1}}$ is a learnable weight matrix at layer $l$, $\tilde{D}$ is the diagonal degree matrix of $\tilde{A}$, and $\sigma$ is an activation function. The term $\tilde{D}^{-\frac{1}{2}} \tilde{A} \tilde{D}^{-\frac{1}{2}} X_t^l W^l$ can be considered as a spatial convolution to aggregate the information from the neighbors which is directly connected.

**ST-GCN block [25].** In general, it is composed of two major blocks: one for spatial convolution and the other for temporal convolution. As is shown in Fig.3, there is a dropout layer between two convolution layers. Both convolution blocks include convolution, BatchNorm (BN) followed by ReLU as the activation function.

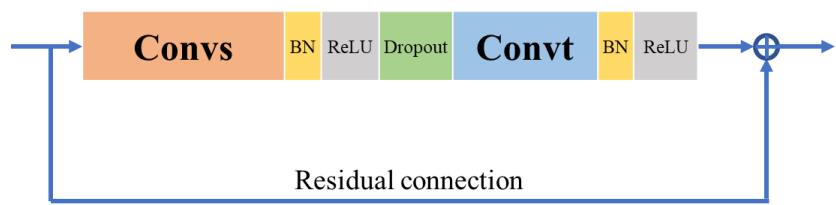

**Fig. 3.** Illustration of the adaptive graph convolutional block. Convs and Convt represent the spatial and temporal GCNs, respectively.

### 3.3.2 Structure of Siamese Networks

**Baseline model.** The network's main architecture adopts AGCN in [24] with adaptions to our use case. The network is composed of ST-GCN blocks, as is shown in Table 2.

**Table 2.** implementation details of the AGCN network

| Layers | input channels | Output channels | Strides | Activation function |
|---|---|---|---|---|
| Batch Norm | - | - | - | ReLU |
| ST-GCN.1 | 3 | 64 | 1 | ReLU |
| ST-GCN.2 | 64 | 64 | 1 | ReLU |
| ST-GCN.3 | 64 | 64 | 1 | ReLU |
| ST-GCN.4 | 64 | 128 | 2 | ReLU |
| ST-GCN.5 | 128 | 128 | 1 | ReLU |
| ST-GCN.6 | 128 | 128 | 1 | ReLU |
| ST-GCN.7 | 128 | 256 | 2 | ReLU |
| ST-GCN.8 | 256 | 256 | 1 | ReLU |
| FC | 256 | 256* | - | - |

*: the channels of FC depend on the shape of embeddings.

**Siamesed networks:** Two copies of augmented features are fed into our model for network training. The siamese network is designed to measure the similarity of gait pairs. The features with the same ID are mapped close to each other while two features with different IDs are mapped far from each other in the feature space.

### 3.3.3 Supervised contrastive loss

**Loss function.** For a graph embedding problem, normalized embeddings from the same class should be pulled closer together than embeddings from different classes.

We take supervised contrastive (SupConLoss), a loss for supervised learning that builds on the contrastive self-supervised literature by leveraging label information [27], as the loss function. The technical novelty in this loss is to consider many positives per anchor in addition to many negatives rather than using one positive (triplet [28]) or one negative (N-pair [29]). The use of many positives and many negatives for each anchor allows us to achieve better performance without the need for hard negative mining, which can be difficult to tune properly. The loss takes features (L2 normalized) and labels as input and shows benefits for robustness to natural corruptions and is more stable to hyperparameter settings such as optimizers and data augmentations [27].

### 3.4 Matching module

For such a large database, the serious performance bottleneck occurs when using the method with high computational complexity. Thus, our matching module calculates the Euclidean distance of the corresponding feature vectors and take the closest sample as the predicted label.

## 4 Experiments

In this part, firstly, we evaluated the recognition accuracy of our method using the rank-1 recognition rates on two datasets: OUMVLP-Pose and CASIA-B. Secondly, we do ablation study on OUMVLP-Pose using different combinations of settings such as different features and different size of input. Thirdly, spatial-temporal study is carried out on OUMVLP-Pose to evaluate the ability of our method to model temporal features.

### 4.1 Dataset and Training Details

Most gait datasets do not provide pose sequence. However, last year, the OU-ISIR Gait Database [9] provided a Multi-View Large Population Database with Pose Sequence (OUMVLP-Pose) to aid research efforts in the general area of developing, testing, and evaluating algorithms for model-based gait recognition.

**Dataset**. OUMVLP-Pose set was built upon OU-MVLP. It contains 10,307 subjects of round-trip walking sequences captured by seven network cameras at intervals of 15° (these sums to 14 views by considering the round trip on the same walking course) with an image size of 1,280 x 980 pixels and a frame rate of 25 fps. Two data sets are provided for OUMVLP-Pose, which were obtained by the OpenPose and AlphaPose models, respectively. These two datasets contain the same number of subjects and the same parameters. Each of the data sets is divided into two disjoint subsets, *i.e.*, training and testing sets with almost the same number of samples [9].

**Protocol**. For a fair comparison, this paper follows the protocol by the OUMVLP-Pose dataset [9]. We divided 10,307 subjects into two sets. The first one with 5,153 subjects is the training set, and the second with the remaining 5,154 subjects is the test set. The test set is separated into a gallery set and a probe set. In our training phase, we set the training batch as 512.

**Details of experiment settings**. In our experiments, we use OneCycle [30] scheduler to dynamically adjust the learning rate and we set the initial learning rate as 0.001. The method is implemented by PyTorch on an RTX 3090 GPU.

**Metric.** The rank-1 recognition rate was employed to evaluate the recognition accuracies.

## 4.2 Performance of networks

First, we evaluated the recognition accuracy of our method with the rank-1 recognition rates on OpenPose and AlphaPose datasets. The results of the 0-270° gallery *vs.* the 0-270° probe are shown in Table 3 and Table 4. It should be mentioned that the horizontal axis represents gallery angle while the vertical axis represents probe angle. The models of both were trained for 1000 epochs. We find that the recognition rate will be relatively high when the probe angle is the same as the gallery angle. The average rate of our method on OpenPose is 38.07% and better than the result 30.80% from [7]. In Table 4, the average rate of AlphaPose (58.02%) is also greater than the result (37.22%) from [7].

**Table 3.** Rank-1 recognition rates by AGCN network using OpenPose dataset for all combinations of views.

| Probe\Gallery | 0 | 15 | 30 | 45 | 60 | 75 | 90 | 180 | 195 | 210 | 225 | 240 | 255 | 270 | mean |
|---|---|---|---|---|---|---|---|---|---|---|---|---|---|---|---|
| 0 | 79.9 | 65.2 | 58.9 | 52.7 | 44.6 | 39.4 | 33.6 | 36 | 41.2 | 35.7 | 44.1 | 42.8 | 39.7 | 37.2 | **25.21** |
| 15 | 75.8 | 86.7 | 78.6 | 74 | 64.6 | 57.5 | 48.3 | 41.6 | 55.2 | 46.7 | 58.5 | 55 | 52.4 | 47.7 | **39.02** |
| 30 | 72.2 | 83.6 | 89.1 | 87 | 77.3 | 71.6 | 61.7 | 44.4 | 57.8 | 55.1 | 67.7 | 64.4 | 61.9 | 57.7 | **45.01** |
| 45 | 64.8 | 77.1 | 86.1 | 90.6 | 85.2 | 79.1 | 68.5 | 42.6 | 55.7 | 53.6 | 70.8 | 68 | 65.2 | 62.2 | **48.1** |
| 60 | 49.6 | 59.7 | 67.6 | 74.5 | 87.1 | 72.6 | 63.5 | 35 | 45.9 | 43.9 | 59.5 | 69.3 | 58.2 | 55.4 | **43.49** |
| 75 | 50.1 | 61 | 72.3 | 80.8 | 85.1 | 90.2 | 82.6 | 37.2 | 48.9 | 47.5 | 66 | 67.6 | 69.8 | 68.3 | **44.87** |
| 90 | 43.8 | 51.3 | 62.1 | 70.1 | 73.7 | 82.3 | 85 | 34 | 44 | 44.1 | 60.4 | 62.4 | 67.6 | 68.7 | **36.16** |
| 180 | 43.1 | 41.6 | 41.8 | 42.6 | 38.1 | 35.8 | 33.2 | 64 | 54.1 | 40.8 | 45.7 | 41.1 | 37.8 | 32.7 | **27.65** |
| 195 | 46.2 | 53 | 52.2 | 51.6 | 45.9 | 43 | 38.2 | 52 | 76.9 | 57.8 | 63.1 | 55.3 | 48.8 | 41.4 | **35.1** |
| 210 | 43 | 48.6 | 52.7 | 52.5 | 46.9 | 44.2 | 40.9 | 41.8 | 61.3 | 65.1 | 68.2 | 58.7 | 53.2 | 46.1 | **34.49** |
| 225 | 49.8 | 56 | 61.8 | 65.8 | 61.2 | 58.6 | 53.7 | 43.5 | 63.1 | 64.4 | 87.1 | 79.5 | 72.6 | 61.9 | **42.37** |
| 240 | 42 | 46.6 | 51.2 | 56.1 | 63.6 | 52.4 | 47.7 | 33.6 | 49.4 | 49.8 | 69.2 | 83.5 | 67.3 | 58.6 | **38.22** |
| 255 | 44.5 | 49.4 | 57 | 62.4 | 61.2 | 63.7 | 60.6 | 34.2 | 49.6 | 50.8 | 74.4 | 79.2 | 86.2 | 78.6 | **39.79** |
| 270 | 42.3 | 45.7 | 53.3 | 57.9 | 58.4 | 62.2 | 62.1 | 30.5 | 43.2 | 44.8 | 65.1 | 68.9 | 78 | 83.3 | **33.53** |
| **mean** | **28.95** | **37.91** | **41.67** | **45.67** | **45.02** | **41.82** | **33.17** | **25.64** | **36.14** | **33.37** | **44.69** | **43.99** | **40.55** | **34.41** | **38.07** |

**Table 4.** Rank-1 recognition rates by AGCN network using AlphaPose dataset for all combinations of views.

| Probe\Gallery | 0 | 15 | 30 | 45 | 60 | 75 | 90 | 180 | 195 | 210 | 225 | 240 | 255 | 270 | mean |
|---|---|---|---|---|---|---|---|---|---|---|---|---|---|---|---|
| 0 | 62.8 | 41.7 | 35.3 | 30.7 | 25.8 | 21 | 14.9 | 16.5 | 18.8 | 15.7 | 19.9 | 18.8 | 17.1 | 13.9 | **46.5** |
| 15 | 52.7 | 77.2 | 61.2 | 56.3 | 47 | 37.1 | 24.9 | 20.2 | 32.6 | 25 | 31.7 | 31.3 | 26.6 | 22.5 | **60.2** |
| 30 | 44.7 | 65 | 78.6 | 74 | 60.8 | 51 | 35.7 | 21.3 | 32.3 | 30.3 | 39 | 36.1 | 33.2 | 28 | **67.97** |
| 45 | 38.6 | 58.8 | 73 | 84.7 | 75.3 | 63.4 | 43.6 | 20.5 | 30.5 | 29.6 | 44.2 | 41 | 37.8 | 32.4 | **69.25** |
| 60 | 29.7 | 44.8 | 52.8 | 66.6 | 82.3 | 63 | 43.7 | 18.7 | 26.9 | 24.5 | 38.4 | 47.7 | 36.6 | 33.2 | **60.13** |
| 75 | 27 | 39.1 | 50.2 | 63.9 | 71.9 | 81.3 | 61.5 | 18.5 | 25.2 | 25.4 | 39.7 | 41.2 | 43.1 | 40.3 | **66.25** |
| 90 | 19.6 | 27.8 | 35.1 | 45.1 | 49.8 | 62.6 | 69.6 | 14.8 | 19.4 | 20 | 31.5 | 33.1 | 38.2 | 39.6 | **60.68** |
| 180 | 20.2 | 21 | 22.2 | 22.3 | 19.9 | 18.4 | 14.5 | 63.1 | 46 | 33.2 | 36.2 | 28.4 | 23.6 | 18.1 | **42.32** |
| 195 | 20.6 | 29.6 | 28.6 | 28.1 | 26.7 | 23.4 | 16.9 | 43 | 74.4 | 49.7 | 51.8 | 41.8 | 32.9 | 23.9 | **51.82** |
| 210 | 17.3 | 24.6 | 28.4 | 29.6 | 26.3 | 23.4 | 18.9 | 32.8 | 52.7 | 61.4 | 59.2 | 44.1 | 36.8 | 27.1 | **51.66** |
| 225 | 20.3 | 29 | 33.4 | 39 | 36.1 | 32.7 | 26.3 | 30.4 | 51.9 | 55.1 | 79.6 | 66.8 | 52.9 | 39.6 | **62.79** |
| 240 | 17.8 | 25.4 | 29.6 | 33.8 | 41.2 | 33 | 26.4 | 22.5 | 37.7 | 38.1 | 59.4 | 76.5 | 54.6 | 39 | **55.06** |
| 255 | 18 | 25.8 | 30 | 34.5 | 35.9 | 39.3 | 32.6 | 20.7 | 33.9 | 33.8 | 55.7 | 63.7 | 75.7 | 57.4 | **60.83** |
| 270 | 15.9 | 21 | 25.1 | 30.6 | 31.3 | 35.8 | 34.9 | 15.9 | 23.7 | 25.4 | 39.3 | 45.4 | 58.6 | 66.5 | **56.83** |
| **mean** | **53.37** | **58.95** | **63.21** | **65.62** | **63.79** | **60.9** | **55.69** | **40.76** | **53.31** | **50** | **64.26** | **63.99** | **61.33** | **57.12** | **58.02** |

### 4.3 Comparison with other model-based methods

Table 5 shows the rank-1 recognition rates of two other model-based methods on the OpenPose and AlphaPose datasets where the probe angle is the same as the gallery angle. One is a recent deep learning-based approach named pose-based temporal-spatial network (PTSN) [20], and the other is the CNN-Pose method [9]. Taking the pose sequences as input, CNN-Pose applies two-dimensional convolution layers, pooling layers, and a full connection layer. For feature extraction, cross-entropy loss based on softmax and a center loss is employed to optimize the network. Wang et al. [7] is the latest SOTA method on OUMVLP-Pose dataset, and it also uses graph convolutional network to embed pose sequence with a simple and naïve future extraction module.

Table 6 shows the average rank-1 recognition rates on different probe angles with three different specific gallery angles. Our method has better performance than the PTSN and CNN-Pose methods [9].

**Table 5.** The rank-1 recognition rates of three model-based methods on the OpenPose and AlphaPose dataset where the probe angle is the same as the gallery angle.

| Methods | 0 | 30 | 60 | 90 | mean |
|---|---|---|---|---|---|
| CNN-Pose (OpenPose) [9] | 31.98 | 53.39 | 69.07 | 37.93 | 48.09 |
| Wang Z et al. (OpenPose) [7] | 56.43 | 73.32 | 77.21 | 63.53 | 67.62 |
| Our method (OpenPose) | **62.84** | **78.64** | **82.28** | **69.55** | **73.33** |
| CNN-Pose (AlphaPose) [9] | 47.25 | 69.13 | 73.21 | 49.07 | 59.67 |
| Wang Z et al. (AlphaPose) [7] | 70.26 | 84.49 | 81.25 | 78.34 | 78.59 |
| Our method (AlphaPose) | **79.54** | **88.43** | **87.08** | **85.15** | **85.50** |

Table 6. The rates are the averages on different probe angles with a specific gallery angle 0°,30°,60°, and 90° on the OpenPose and AlphaPose datasets.

| Methods | 0 | 30 | 60 | 90 | mean |
|---|---|---|---|---|---|
| CNN-Pose (OpenPose) [9] | 7.52 | 18.56 | 22.81 | 11.83 | 15.18 |
| Wang Z et al. (OpenPose) [7] | 22.81 | 29.78 | 32.74 | **36.32** | 29.23 |
| Our method (OpenPose) | **28.95** | **41.67** | **45.02** | 33.17 | **37.20** |
| CNN-Pose (AlphaPose) [9] | 12.27 | 29.32 | 30.50 | 18.06 | 22.54 |
| Wang Z et al. (AlphaPose) [7] | 32.21 | 38.11 | 41.49 | 43.18 | 38.32 |
| Our method (AlphaPose) | **40.96** | **50.99** | **51.76** | **44.16** | **46.97** |

### 4.4 Result on CASIA-B dataset and comparison with appearance-based method

CASIA-B [7] is a widely used gait dataset composed of 124 subjects. For each subject, the dataset contains 11 views (0°, 18°, . . ., 180°) and 3 waking conditions. The walking conditions are normal (NM), walking with a bag (BG), and wearing a coat or a jacket (CL). For getting pose sequences from the raw input images, we use OpenPose [31] as a 2D human pose estimator to obtain the human pose in each frame. Table 7 shows the comparison of our method with GaitNet [1], which is a typical appearance-based approach to gait recognition in recent years. Notably, with our lower dimension feature representation, we can still achieve competitive results against the appearance-based method.

Table 7. Rank-1 recognition rates by AGCN network using CASIA-B dataset with OpenPose for all combinations of views.

| | 0 | 18 | 36 | 54 | 72 | 90 | 108 | 126 | 144 | 162 | 180 | mean | GaitNet[21] |
|---|---|---|---|---|---|---|---|---|---|---|---|---|---|
| NM#5-6 | 72.4 | 81.2 | 85.6 | 80.4 | 79.4 | 85.0 | 81.0 | 77.6 | 82.5 | 79.1 | 80.2 | 80.4 | 91.6 |
| BG#1-2 | 62.5 | 68.7 | 69.4 | 64.8 | 62.8 | 67.2 | 68.3 | 65.7 | 60.7 | 64.1 | 60.3 | 65.0 | 85.7 |
| CL#1-2 | 57.8 | 63.2 | 68.3 | 64.1 | 66.0 | 64.8 | 67.7 | 60.2 | 66.0 | 68.3 | 60.3 | 64.2 | 58.9 |

### 4.5 Ablation Study

To verify the effectiveness of our method, the ablation experiments are given in this section, the experiments are conducted on the OUMVLP-Pose dataset using different combinations of settings. Due to the limitation of time and sources, the following results are all trained for 500 epochs.

**Impact of samples' quantity.** The performance of our method on small dataset is evaluated and shown in Table 8. In this experiment, we feed 20%, 40%, 60%, 80% of the train set to different model. Now, using only 40% of the data can achieve the results in Wang Z. et al [7], which shows the power of our network.

**Table 8.** Rank-1 recognition mean rates using AlphaPose with different percent of training datasets.

|  | 20% | 40% | 60% | 80% | 100% |
|---|---|---|---|---|---|
| Our method (500 epochs) | 27.13 | 41.10 | 45.16 | 48.67 | 51.97 |
| Wang et al [7] | 12.90 | 21.89 | 28.96 | 35.22 | 37.22 |

**Impact of local features.**

As shown in Fig 4 and Fig 5, we have compared the recognition accuracies using all local features and isolated local features. It can be found that: 1) The performances of using isolate local features are no better than using all local features, which is reasonable since all local features utilize more information for recognition; 2) The performance of joint features is better than bone features for AlphaPose dataset, while that of the bone features is better than joint features for OpenPose dataset. The possible reason is that the coordinates of a large number of joints are estimated as 0 in the OpenPose dataset. Since the joint feature directly uses the joint point coordinates, it is more likely to be disturbed by invalid data. However, the bone feature describes the information of connected joints, which reduces the interference of invalid data.

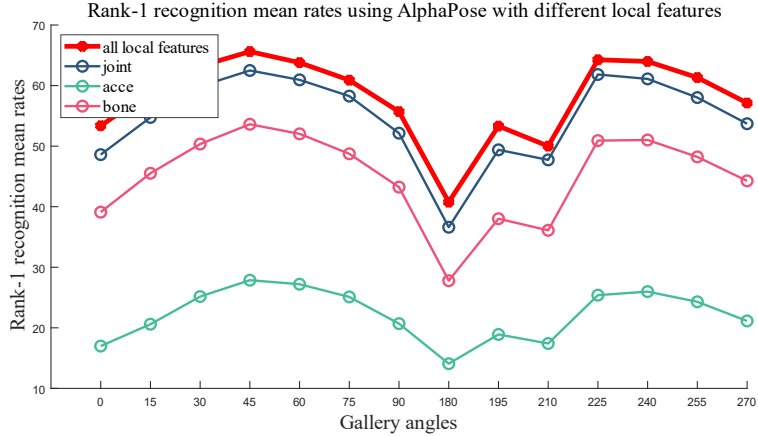

**Fig. 4.** Rank-1 recognition mean rates using AlphaPose with different local features

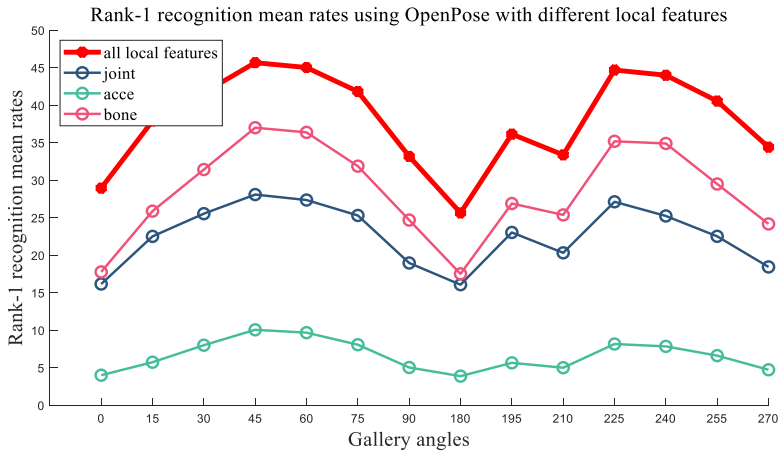

**Fig. 5.** Rank-1 recognition mean rates using OpenPose with different local features

### 4.6 Spatio-temporal Study.

Furthermore, our approach shows a high ability to model temporal features as shown in Table 9. When trained with sorted sequences and tested with shuffled sequences (row 3), the performance drops significantly. These results support that our method relies on temporal features within pose sequences. Table 9 also illustrates the spatial modeling abilities in row 1. Despite the missing temporal and appearance information, the network is still able to learn spatial features of gait.

**Table 9.** Control Condition: sort/shuffle the input sequence at the train/test phase. Results are rank-1 accuracies averaged on 14 views.

|   | Train   | Test    | OpenPose | AlphaPose |
|---|---------|---------|----------|-----------|
| 1 | Shuffle | Sort    | 23.64    | 37.98     |
| 2 | Sort    | Sort    | 38.07    | 58.02     |
| 3 | Sort    | Shuffle | 10.58    | 22.01     |

## 5 Conclusion

In this work, we propose a model-based gait recognition approach by adopting siamese structure and taking the AGCN network as the embedding network. This model-based approach increases the flexibility of the graph convolutional network and is more robust to the gait recognition task in the real world. Furthermore, we build the joint, the acceleration, and bone's angle features to enrich the input of the network. The traditional methods always ignore or underestimate the importance of the skeleton information because it used to be hard to fit the human model (or estimate human pose). The proposed model is evaluated on CASIA-B and OUMVLP-Pose, a large population database with multi-view pose sequences, and achieves the new state-of-the-art performance.

**Acknowledgement** This work was supported by the Key Research & Development Programs of Jiangsu Province (BE2018720) and the Open project of Engineering Center of Ministry of Education (NJ2020004).